\documentclass[twoside]{IEEEtran}
\ifCLASSINFOpdf
\else
\fi
\usepackage{cite}
\usepackage{amsmath,amssymb,amsfonts}
\usepackage{algorithmic}
\usepackage{graphicx}
\usepackage{subcaption}
\usepackage{textcomp}
\usepackage{wrapfig}
\usepackage{algorithm}
\usepackage{lipsum}
\usepackage[hidelinks]{hyperref}
\usepackage{xcolor}
\usepackage{tabularx,booktabs}
\newcolumntype{C}{>{\centering\arraybackslash}X} % centered version of "X" type
\def\BibTeX{{\rm B\kern-.05em{\sc i\kern-.025em b}\kern-.08em
    T\kern-.1667em\lower.7ex\hbox{E}\kern-.125emX}}
\begin{document}
\title{Continuously Learning to Detect People on the Fly: A Bio-inspired Visual System for Drones} %instead of navigation
%Towards 100 FPS, High-Precision, Fail-Safe Human Detection for Drones Through Curriculum Learning

\author{Ali Safa, \IEEEmembership{Student Member, IEEE}, Ilja Ocket, \IEEEmembership{Member, IEEE}, 
Andr\'e Bourdoux, \IEEEmembership{Senior Member, IEEE}, 
\\
Hichem Sahli,
Francky Catthoor, \IEEEmembership{Fellow, IEEE}, 
Georges G.E. Gielen, \IEEEmembership{Fellow, IEEE}

%\thanks{Manuscript received: July 15, 2021; Revised:
%September 25, 2021; Accepted: November 1, 2021.}
%\thanks{This paper was recommended for publication by Editor Cesar Cadena Lerma upon evaluation of the Associate Editor and Reviewers’ comments.}
\thanks{A. Safa, I. Ocket, F. Catthoor and G. G.E Gielen are with imec and KU Leuven, 3001 Leuven, Belgium (Ali.Safa-Ilja.Ocket-Francky.Catthoor@imec.be; Georges.Gielen@kuleuven.be).  A. Bourdoux is with imec, 3001 Leuven, Belgium (Andre.Bourdoux@imec.be). H. Sahli is with imec and ETRO VUB, 1050 Brussels, Belgium (hsahli@etrovub.be)}% <-this % stops a space
%\thanks{Andr\'e Bourdoux is with imec, 3001 Leuven, Belgium (e-mail: Andre.Bourdoux@imec.be).}
%\thanks{Funding: “Onderzoeksprogramma Artificiële Intelligentie (AI) Vlaanderen”}
%\thanks{The research leading to these results has received funding from the Flemish Government (AI Research Program) and the European Union's ECSEL Joint Undertaking under grant agreement n° 826655 - project TEMPO.}
%\thanks{Digital Object Identifier (DOI): see top of this page.}
}

\markboth{}%
{Safa \MakeLowercase{\textit{et al.}}: }
%IEEE ROBOTICS AND AUTOMATION LETTERS. PREPRINT VERSION. ACCEPTED November, 2021
\markboth{}{Safa \MakeLowercase{\textit{et al.}}: SHORTTITLE}




\maketitle

\begin{abstract}
 This paper demonstrates for the first time that a biologically-plausible spiking neural network (SNN) equipped with Spike-Timing-Dependent Plasticity (STDP) can continuously learn to detect walking people on the fly using retina-inspired, event-based cameras. Our pipeline works as follows. First, a short sequence of event data ($<2$ minutes), capturing a walking human by a flying drone, is forwarded to a convolutional SNN-STDP system which also receives teacher spiking signals from a readout (forming a semi-supervised system). Then, STDP adaptation is stopped and the learned system is assessed on testing sequences. We conduct several experiments to study the effect of key parameters in our system and to compare it against conventionally-trained CNNs. We show that our system reaches a higher peak $F_1$ score (+19\%) compared to CNNs with event-based camera frames, while enabling on-line adaptation.
 
 %we compare our precision-recall performance to conventionally-trained CNNs working with both RGB or event-based camera frames.
\end{abstract}


%\section*{Supplementary Material}
%\noindent
%Please find additional videos at: \url{https://tinyurl.com/4cvhymbp}  

% \fi
\IEEEpeerreviewmaketitle

\section{Introduction}
\label{sec:introduction}
%\textcolor{blue}{annonce le budget power pour apres (pas yolo)}
\IEEEPARstart{I}{n} recent years, the use of micro drones has attracted much attention for applications ranging from infrastructure inspection to people search and rescue \cite{falanga}. In those applications, drones and humans will be moving within the same environment. Therefore, it is critical to equip micro drones with people detection pipelines for safety purposes \cite{ali}. %problem dnn not ultra-low power, small dnn not lot of expressivity, fast adaptation cannot be done since too costly, training in backend may be too slow and violate law in real-world application, backend may fail -> snn solves power, stdp solves adaptation "bio inspired" 

Following the enormous progress in \textit{deep learning}, convolutional neural networks (CNNs) such as \textit{You Only Look Once} (YOLO) and its variants, constitute the state of the art in terms of detection and speed performance \cite{tinyYOLO}. %They reach high frame rates when running on power-hungry, desktop-grade GPUs ($\sim 65$ FPS) \cite{YOLOv4} or on \textit{less} power-hungry \textit{embedded} GPUs such as the NVIDIA Jetson Nano ($\sim 30$ FPS for a Tiny-YOLO \cite{tinyYOLO}). 
%du coup, pas plus mal d'etre dans le server the typical way of doing is server but problem is with server breaks e.g., (encryption ok)
However, the use of such CNNs is not suited for micro drones because of the significant power budget that conventional CNNs demand ($2$W for an edge TPU, $10$W for a Jetson Nano) relative to the micro drone power budget ($<1$W). Therefore, a significant effort has been put in the co-design of CNNs and micro-controller (MCU) architectures in order to reduce the power and memory consumption budget of the vision system down to $<100$mW and $<1$MB respectively \cite{computematter}. Still, adopting a conventional CNN-hardware design approach leads to rigid systems that cannot easily adapt themselves to new environments and tasks (unlike the nervous systems of biological entities in nature). 
%Another promising route towards ultra-low-power and -area perception is to take inspiration from biology. 
%Still, in order to use those DNNs on micro drones, a significant power budget of $\sim 10$W must be spent when using e.g., a Jetson Nano on a micro drone (consuming $10$-$50$W \cite{powmodel}), which strongly limits the drone flight time and payload that can be carried \cite{computematter}. %\textcolor{blue}{This is even more problematic for ultra-small, insect-like drones + price of other sensors} 
%On the other hand, the use of smaller networks running on highly power-optimized DNN processors only partially solve the problem since a lower detection performance is typically reached compared to deeper networks%, due to a decrease in network \textit{expressivity} (e.g., $>50\%$ decrease in \textit{mean average precision} for Tiny-YOLO vs. YOLO) 
%\cite{yolov3}.
\setcounter{figure}{1} 
\begin{figure}[htbp]
\centering
    \includegraphics[scale = 0.36]{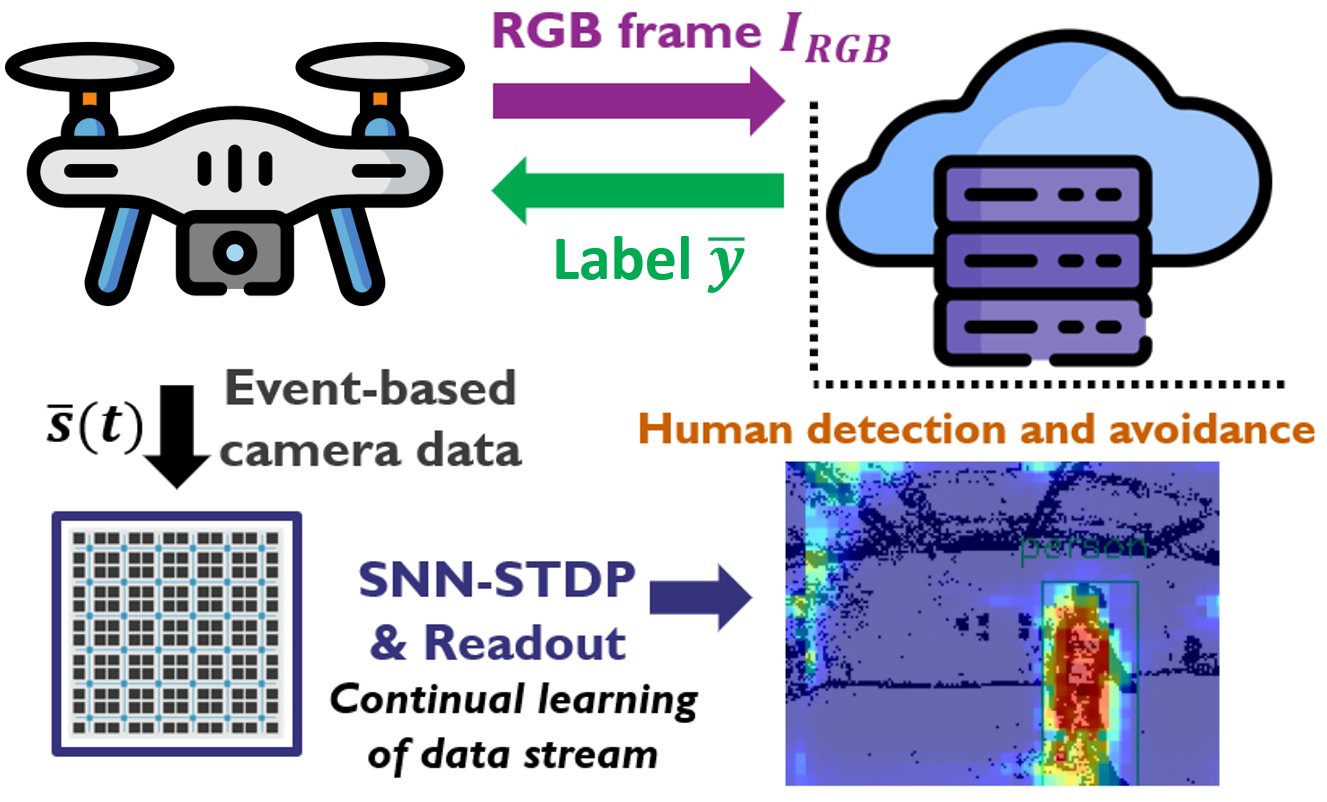}
    \caption{\textit{\textbf{Continual learning of people detection}. Conceptual illustration of a micro-drone equipped with an event-based camera, an RGB camera and a bio-inspired SNN-STDP network. The RGB frames are sent to a back-end server where an \textit{oracle} CNN returns people detection labels. The labels are used during the \textit{continual learning} of the SNN-STDP system%which continuously infers an attention map from the event data
    . %\textcolor{blue}{comment that?} Such low-memory-footprint system could be used to enhance the safety of drones is situations where server latency is too high or in case of a server crash. 
    }}
    \label{scenariodrone1}
\end{figure}

Indeed, conventional CNNs rely on the standard off-line training procedure where the CNN is trained on a large dataset capturing the task that needs to be solved. However, it is not always guaranteed that all scenarios to be encountered at run time are well captured by the training set (e.g., partially damaged infrastructures are hard to capture by a dedicated dataset for \textit{indoor search and rescue} tasks). In addition, since CNN training requires desktop-grade computing, the conventional solution for fine-tuning on-board CNNs is to first re-train the network offline and then send the updated weights back to the drone. Clearly, this conventional approach has many drawbacks such as high latency (training takes a long time, prohibiting fast adaptation) and privacy concerns due to the on-line storage of user data (also causing legal issues due to e.g., the \textit{right to data erasure} in the EU\footnote{\url{https://gdpr.eu/right-to-be-forgotten/}}). %In addition, this approach is very far from biological plausibility.

%for novel environments is i) to continuously acquire raw data, ii) generate labels in the back-end server using an \textit{oracle} system (e.g., high-performance DNN), iii) re-train the DNN in the back-end server with the updated dataset and iv) send the updated weights back to the drone. This conventional approach has many drawbacks. It is very latency-expensive (re-training takes a long time), prohibiting the fast adaptation of the vision system to sudden changes in the environment. In addition, it requires the system to log a significant amount of user data to update the training set, causing security, privacy and legal concerns (e.g., \textit{right to erasure} in the EU\footnote{\url{https://gdpr.eu/right-to-be-forgotten/}}) . %if we want commercially available stuff, uk eu 

Recently, the use of biologically-inspired \textit{spiking neural networks} (SNNs) has gained huge interest for the design of ultra-low-power AI-enabled systems \cite{preprint, yulia}. In contrast to CNNs, SNNs make use of spiking neurons that communicate through low-complexity binary activations in an \textit{event-based} manner (vs. frame-based processing in CNNs), only consuming energy when a spike is emitted \cite{preprint}. In addition, SNNs can be implemented in massively parallel, \textit{non Von Neumann} computing architectures, solving the energy- and latency-expensive memory bottleneck issues \cite{yulia}. Finally, the use of a bio-inspired learning rule (vs. backprop) working in the binary activation domain and \textit{local} to each neuron (such as \textit{Spike-Timing-Dependent Plasticity} or STDP \cite{Bi10464}) enables ultra-low-power learning at the edge (not tackled by embedded CNNs) \cite{charlotte}. Therefore, a growing number of SNN-STDP computing units have been proposed, achieving a power consumption as low as $\sim 10$0$\mu$W \cite{charlotte} while enabling \textit{on-chip inference and learning} (three orders of magnitude lower power consumption vs. optimized CNNs running on MCU \cite{computematter}). 

As SNNs require spiking data as input, they are often used in conjunction with an \textit{event-based camera} (also called \textit{dynamic vision sensor} or DVS), inspired by the inner working of the human eye. DVS cameras are composed of independent pixels that emit spikes asynchronously whenever the change in light log-intensity crosses a threshold $C$ \cite{gallego} (see Fig. \ref{dvsimg}).
%As SNNs require spiking data as input, they are often used in conjunction with an \textit{event-based camera} (also called \textit{dynamic vision sensor} or DVS, see Fig. \ref{dvsimg}). DVS cameras are novel bio-inspired vision sensors composed of independent pixels that asynchronously emit spikes whenever the change in light log-intensity crosses a certain threshold \cite{gallego}. Compared to standard imaging cameras, DVS sensors have a ultra-fine-grain time resolution ($\sim 1\mu$s), a high dynamic range ($\sim 140$dB vs. $\sim 60$dB) and do not suffer from motion blur \cite{gallego}.
\setcounter{figure}{0} 
\begin{figure}[htbp]
\centering
    \includegraphics[scale = 0.53]{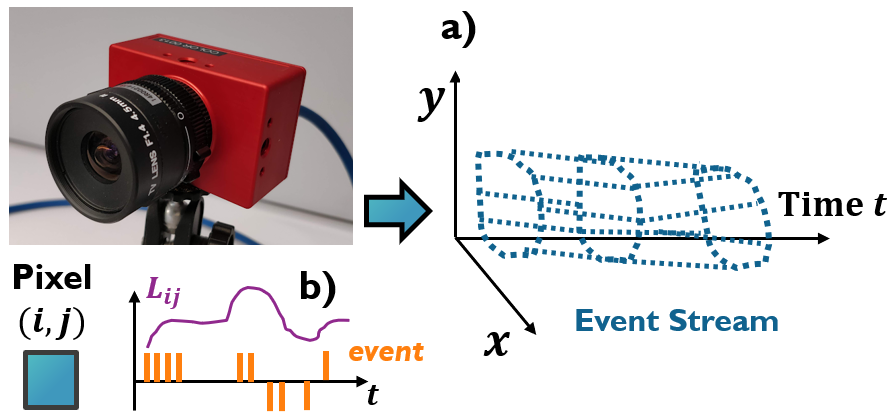}
    \caption{\textit{\textbf{Conceptual illustration of event-based data.} The DVS is capturing an image of "0". a) The DVS outputs a spike stream in time and space. b) When the change $|\Delta L_{ij}|$ in light $L_{ij}$ at pixel $(i,j)$ crosses a threshold $C$, a spike is emitted. The spike is positive if $\Delta L_{ij}>0$ and negative otherwise.}}
    \label{dvsimg}
\end{figure}

In this paper, our goal is to investigate the design of a bio-inspired SNN-STDP system that can \textit{continuously learn} to perform people detection on drones from DVS camera data. Fig. \ref{scenariodrone1} illustrates the scenario that we consider in this work: the drone must explore the environment during $\sim 2$ minutes and continuously learn to detect a walking human subject using the current data sample only, \textit{without access to the past and the future data}. Then, continual learning (CL) is stopped and we assess the detection performance on a testing sequence. %The main contributions of this work are as follows. We propose a semi-supervised SNN-STDP framework 

As the basis for our people detection network, we use the state-of-the-art SNN-STDP architecture proposed in \cite{preprint} and we augment it with three significant enhancements: \textit{i)} we motivate and extend the SNN-STDP network to the use of \textit{anti-Hebbian} (negative STDP) learning rules; \textit{ii)} we propose a semi-supervised method that allows the learning of the logistic regression readout and the SNN \textit{at the same time} (vs. disjoint learning in \cite{preprint}) and \textit{iii)} we use our SNN in a more challenging CL setting (vs. offline training in \cite{preprint}).

This paper is organized as follows. Related works are discussed in Section \ref{related}. Background theory is given in Section \ref{background}. Our methods are presented in Section \ref{methods}. Results are discussed in Section \ref{expres}. Conclusions are given in Section \ref{conc}. 

% \subsection*{Aim of this work}
% In this paper, our goal is to investigate the design of a \textit{continual learning} (CL) system built around a DVS camera and a convolutional SNN architecture, for the task of people detection on micro drones. Fig. \ref{scenariodrone1} illustrates the scenario that we consider in this paper: the drone must explore the environment during $\sim 2$ minutes and continuously learn to detect a walking human subject \textit{without access to the past and the future data} (no shuffled dataset and off-line training). Then, continual learning is stopped and we assess the detection performance of the vision system on separate testing sequences. 

% Our work significantly deviates from previously proposed vision pipelines for drones by presenting a \textit{biologically-plausible} visual system incorporating CL via STDP, as empirically observed in the brain \cite{Bi10464}, that can learn a given environment using a very limited amount of data. Therefore, the main contribution of this work is to debut what is, to the best of our knowledge, the first study of STDP-based bio-inspired drone vision \textit{that can learn on the fly}. This paper is organized as follows. Related works are discussed in Section \ref{related}. Background theory is given in Section \ref{background}. Our methods are presented in Section \ref{methods}. Experimental results are discussed in Section \ref{expres}. Conclusions are provided in Seection \ref{conc}. 
\setcounter{figure}{3} 
\begin{figure*}[htbp]
\centering
%\captionsetup{justification=centering}
    \includegraphics[scale = 0.49]{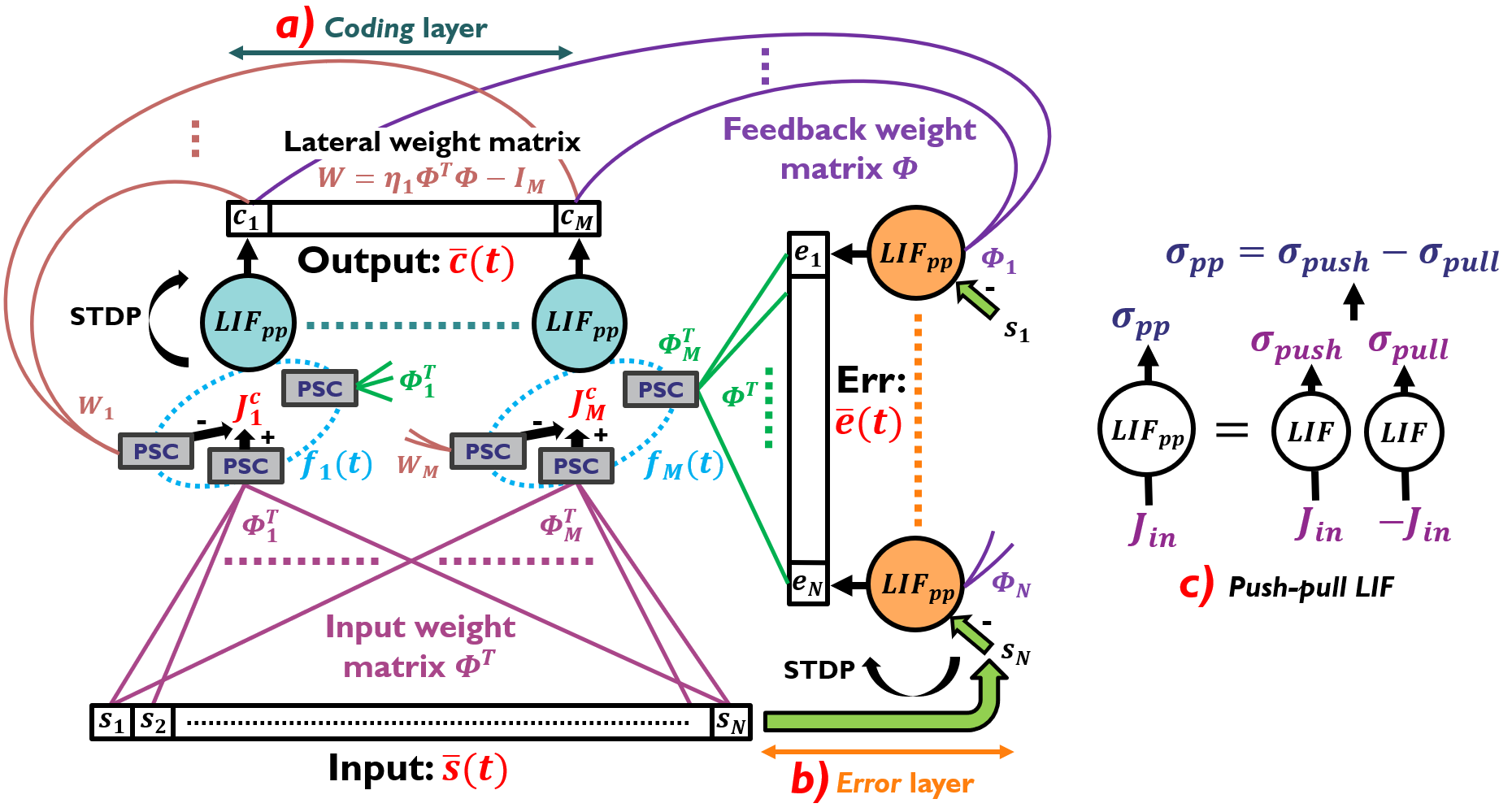}
    \caption{\textit{\textbf{SNN-STDP architecture.} a) The coding layer features $i=1,...,M$ pairs of push-pull LIF neurons ($LIF_{pp}$) that receive the spiking vector $\bar{s}$ (input) through $\Phi^T$ (input weights) and the previous spiking output vector $\bar{c}$ through $W$ (lateral weights). The output spike trains $\bar{c}$ are obtained following (\ref{infer}). In addition, the coding layer receives the re-projection error spike trains $\bar{e}$ through $\Phi^T$ and combines them with the other inputs using (\ref{waytocompute}) to obtain $f_i$ locally to each coding neuron. STDP is applied for learning $\Phi^T$ and $W$ with $c_i$ and $f_i$ the post-synaptic and $\bar{e}$ the pre-synaptic spike trains. b) The error layer features $j = 1,...,N$ pairs of push-pull LIF neurons $LIF_{pp}$ which receive the entry $s_j$ as input and the SNN output $\bar{c}$ through $\Phi$ (feedback weights, remain identical to the transpose of input weights during learning). The error layer outputs the spiking vector $\bar{e}$ following (\ref{eerr}). STDP is applied for learning $\Phi$ with each $e_j(t)$ as the post-synaptic spike train and $\bar{c}$ as the pre-synaptic spike trains. c) The pair of push-pull LIF neurons codes both positive and negative values of the input current $J_{in}$ (output spikes of positive or negative polarity).}}
    \label{fig1}
\end{figure*}

% the right to be forgotten of EU GDPR
% 2Art. 17 GDPR – Right to erasure (right to be forgotten)
% EU: \url{https://gdpr.eu/right-to-be-forgotten/}
% UK:\url{https://ico.org.uk/for-organisations/guide-to-data-protection/guide-to-the-general-data-protection-regulation-gdpr/individual-rights/right-to-erasure/}
\section{Related Works}
\label{related}
A growing number of bio-inspired robotic systems have been proposed in the past decades, mainly focusing on the study of animal-like actuation and motor control \cite{auke}. Complementary to bio-actuation, a number of perception systems for drones, taking inspiration from the inner workings of the human eye and the visual cortex, have recently emerged thanks to the advent of neuromorphic event-based cameras \cite{yulia, droneavoid, gallego}. A line tracking system for drones was proposed in \cite{yulia} using a DVS camera processed by an SNN. Similar to our work, the system of \cite{yulia} uses an SNN with local bio-inspired learning rules (analogous to STDP) and demonstrates that such adaptive SNN system can learn to compensate for external disturbances online. Closer to our work, a DVS-based moving object tracking system for drones was proposed in \cite{droneavoid}, using hand-crafted feature extraction followed by Kalman filtering (processing pipeline not bio-plausible). Compared to the hand-crafted feature extraction in \cite{droneavoid}, we proposed a bio-inspired vision system that can continuously \textit{learn} to detect walking people from a drone (vs. detecting \textit{any} moving object in \cite{droneavoid}). 

Since sparsely-supervised and unsupervised learning are believed to be key mechanisms in the brain, a number of bio-inspired architectures have been proposed for learning spatio-temporal features from event-based data \cite{hots, preprint}. Among them, the \textit{hierarchy of event-based time surfaces} (HOTS) was proposed in \cite{hots} by cascading unsupervised layers, trained to cluster event-data aggregated as \textit{time surfaces} (exponential decay maps) and by using an output readout (e.g., logistic regression) to classify the extracted features. In contrast to HOTS, which uses conventional clustering methods, an unsupervised SNN-STDP network with better biological plausibility was proposed by us in \cite{preprint} for DVS feature learning, achieving state-of-the-art performance on common DVS benchmarks. Thus, we use the network of \cite{preprint} as our basis in this work.  

%In this paper, we use the SNN-STDP architecture proposed in \cite{preprint} as the basis for our people detection network. We introduce three significant enhancements to the architecture of \cite{preprint}: \textit{i)} we motivate and extend the SNN-STDP network to the use of \textit{anti-Hebbian} (negative STDP) learning rules; \textit{ii)} we propose a semi-supervised method that allows the learning of the readout and the SNN \textit{at the same time} (vs. disjoint learning in \cite{preprint, hots}) and \textit{iii)} we use our SNN in a more challenging continual learning setting (vs. offline training in \cite{preprint}).

\section{Background theory}
\label{background}
\subsection{SNN-STDP fundamentals}
In contrast to CNNs, SNNs make use of \textit{spiking} neurons, often modelled by a Leaky Integrate-and-Fire (LIF) activation:
\begin{equation}
 \begin{cases}
    \frac{dV}{dt} = \frac{1}{\tau_m} (J_{in} - V)
    \\
    \sigma = 1, V \xleftarrow{} 0 \hspace{3pt} \text{\textbf{if}} \hspace{3pt} V \geq \mu, \hspace{3pt} \text{\textbf{else}} \hspace{3pt} \sigma = 0
  \end{cases}
  \label{liff}
\end{equation}
with $J_{in}$ the input current to the neuron, $\sigma$ the spiking output, $V$ the membrane potential, $\tau_m$ the membrane time constant and $\mu$ the neuron threshold \cite{preprint}. The scalar input current $J_{in}$ is continuously integrated in $V$ following (\ref{liff}). When $V$ crosses the firing threshold $\mu$, the membrane potential is reset back to zero and an output spike is emitted. The input current $J_{in}$ is obtained by filtering the inner product of the neural weights and the spiking inputs through a \textit{post-synaptic current} (PSC) kernel \cite{preprint} (estimating the spiking \textit{rate}):
\begin{equation}
    J_{in} = \mathcal{PSC}\{ \Bar{w}^T \Bar{s}_{in}(t)  \}
    \label{inee}
\end{equation}
with $\Bar{s}_{in}(t)$ the input spiking vector (originating from e.g. an event camera or other spiking neurons), $\Bar{w}$ the weight vector and $\mathcal{P}\mathcal{S}\mathcal{C} \{ \theta(t) \} = \theta(t) * \frac{1}{\tau_s} e^{-t/\tau_s}$
% \begin{equation}
%     \mathcal{P}\mathcal{S}\mathcal{C} \{ \theta(t) \} = \theta(t) * \frac{1}{\tau_s} e^{-t/\tau_s}
% \end{equation}
the effect of PSC filtering with time constant $\tau_s$. Fig. \ref{lifconcept} a) conceptually illustrates the LIF neuron behaviour.
\setcounter{figure}{2} 
\begin{figure}[htbp]
\centering
    \includegraphics[scale = 0.55]{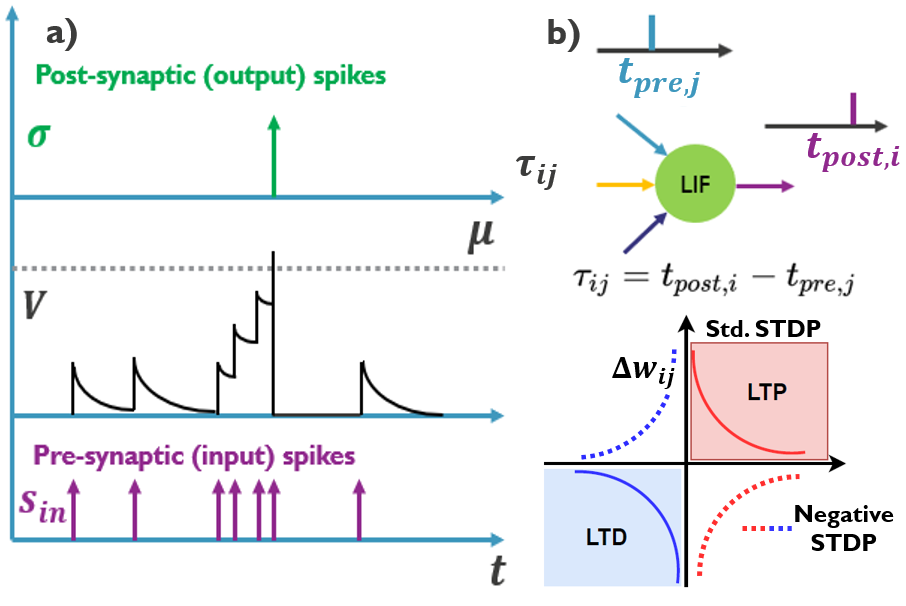}
    \caption{\textit{\textbf{Conceptual illustration of the LIF neuron and STDP learning}. a) The LIF receives as input a pre-synaptic spike train $s_{in}$. When the LIF membrane potential $V$ rises above the threshold $\mu$, a spike is emitted ($\sigma = 1$) and $V$ is set back to zero. b) The synapse strength is modified following the double exponential STDP rule (\ref{stdppp}) in function of the difference between the post- and the pre-synaptic spike times $\tau_{ij} = t_{post,i} - t_{pre,j}$.}}
    \label{lifconcept}
\end{figure}
In the SNN architecture used in this work, learning is performed using the STDP rule, \textit{locally} modifying the weights $w_{ij}$ of each neuron $i$ as follows:
\begin{equation}
   w_{ij} \xleftarrow{} \begin{cases}
  w_{ij} + A_{+} e^{-\tau_{ij}/ \tau_+}, & \text{if } \tau_{ij} \geq 0
\\
   w_{ij} -A_{-} e^{\tau_{ij}/ \tau_-}, & \text{if } \tau_{ij} < 0
\end{cases}
\label{stdppp}
\end{equation}
with $A_+, A_-$ the long-term potentiation (LTP) and depression (LTD) weights, $\tau_+, \tau_-$ the potentiation and depression decay constants, $w_{ij}$ the $j^{th}$ element of the $i^{th}$ neuron weight vector $\Bar{w}$ and $\tau_{ij}$ the time difference between the post- and the pre-synaptic spike times across the $j^{th}$ synapse of neuron $i$ (see Fig. \ref{lifconcept} b) \cite{Bi10464}. 
Finally, it can be shown \cite{preprint} that a good approximation for the long-term effect of STDP (i.e., its expected value over time $t$) is given by the product of the post- and pre-synaptic \textit{mean spike rates} (noted $r_{post,pre}$):
\begin{equation}
    \Delta w_{\text{STDP}} \{ s_{post}, s_{pre}\} \approx \eta_2 (A_+ \tau_+ - A_- \tau_-)r_{post} r_{pre}
    \label{approx}
\end{equation}
where $\eta_2$ is the learning rate.
\subsection{SNN-STDP as Dictionary Learning, Basis Pursuit (DLBP)}
As proposed in \cite{preprint}, the SNN-STDP in Fig. \ref{fig1} is \textit{iteratively} solving the \textit{unsupervised} DLBP problem (inferring $\bar{c}$ using the current $\Phi$ and learning the next $\Phi$ using the current $\bar{c}$):
\begin{equation}
    \Bar{c}, \Phi = \arg \min_{\Bar{c}, \Phi} \frac{1}{2} ||\Phi \Bar{c} - \Bar{s}||_{2}^{2} + \lambda_1 ||\Bar{c}||_1 + \frac{\lambda_2}{2} ||\Phi||_F^2
    \label{lassodefdivc}
\end{equation}
where $\bar{c}$ is the $M$-dimensional output vector containing the mean spike rates of each output neuron in the SNN, $\Phi$ is the learned network weight matrix, $\bar{s}$ is the $N$-dimensional input vector, $\lambda_1$ is a sparsity-controlling hyper-parameter and $\lambda_2$ is a weight decay hyper-parameter.
\setcounter{figure}{4} 
\begin{figure*}[htbp]
\centering
%\captionsetup{justification=centering}
    \includegraphics[scale = 0.57]{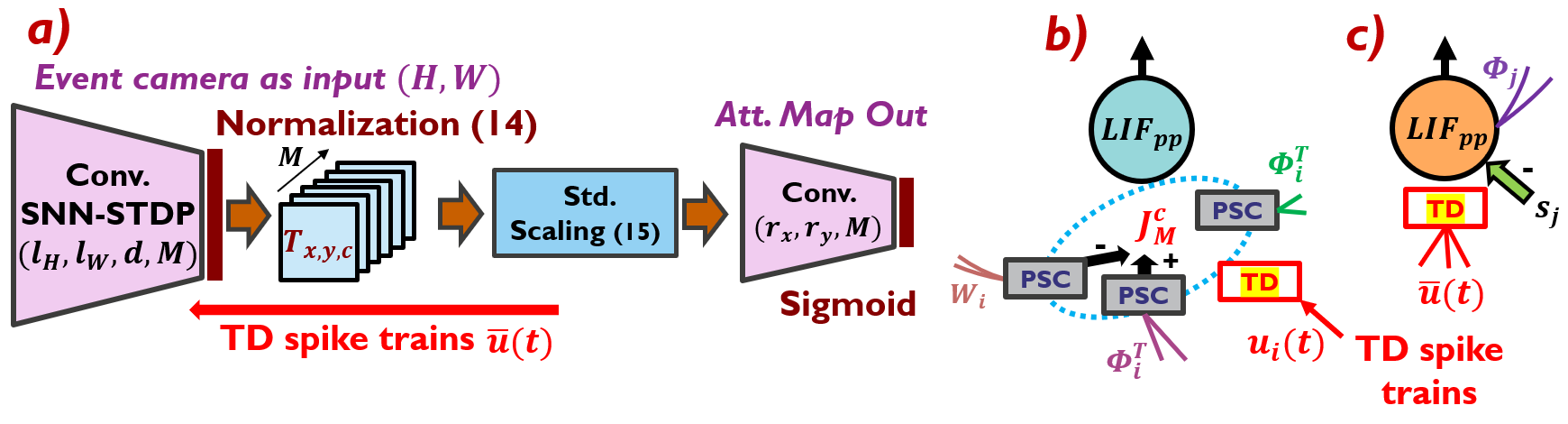}
    \caption{\textit{\textbf{Network architecture for CL and people detection using a DVS camera}. a) Convolutional SNN-STDP followed by a convolutional readout layer. The SNN output is normalized using (\ref{norm}). The normalized tensor is then fed to a standard scaling block (\ref{norm2}) and the resulting tensor is processed by the readout. The readout outputs an attention map indicating the position of a human subject on the image plane. The readout continuously outputs a task-driven (TD) signal $\bar{u}(t)$ following (\ref{epsig}). b) The $i^{th}$ neuron in the coding layer (see Fig. \ref{fig1} a) is augmented with an additional synapse to receive the $i^{th}$ component of the TD signal $u_i(t)$. c) The $j^{th}$ neuron in the error layer (see Fig. \ref{fig1} b) is augmented with additional inputs receiving $\bar{u}(t)$.} }
    \label{figarch2}
\end{figure*}

DLBP (\ref{lassodefdivc}) is solved iteratively by applying STDP across all neurons as follows \cite{preprint}. The $j^{th}$ neuron in the error layer (see Fig. \ref{fig1} b) receives as input the spike trains from the coding layer and outputs the reconstruction error in the spike domain:
\begin{equation}
    e_j = LIF_{pp}\{\Phi_j \bar{c} - s_j\}
    \label{eerr}
\end{equation}
where $\Phi_j$ is locally modified by applying STDP with the $e_j$ as the post- and $\bar{c}(t)$ as the pre-synaptic spike trains. $LIF_{pp}$ denotes the application of a push-pull pair of LIF neurons (\ref{liff}) (see Fig. \ref{fig1} c). The $i^{th}$ neuron in the coding layer (see Fig. \ref{fig1} a) receives $\bar{s}$ and the \textit{previous} output vector $\bar{c}$ as input, and generates the output (where $\Phi_i,W_i$ are the $i^{th}$ matrix rows):
\begin{equation}
    c_i = LIF_{pp}\{\Phi_i \bar{s} - W_i \bar{c}\} \hspace{3pt} \textbf{,} \hspace{3pt} W \equiv \eta_1 \Tilde{W} - I_M \hspace{3pt} \textbf{,} \hspace{3pt}  \Tilde{W} \equiv \Phi^T \Phi
    \label{infer}
\end{equation}
with $\eta_1$ the output convergence rate (to be set) and $I_M$ the identity matrix of dimension $M$. Each coding neuron $i$ also receives the error vector $\bar{e}$ from the error layer and locally combines it with its inputs $\bar{c}$ and $\bar{s}$ to form the local signal $f_i$:
\begin{multline}
   f_i(t) =  \underbrace{\eta_1^{-1}(W + I_M)_i}_{\Tilde{W}_i \hspace{3pt}  (\ref{infer})}\Bar{c}(t) - (\Phi^T)_i \Bar{e}(t) - (\Phi^T)_i \Bar{s}(t) \\[-15pt]
  \stackrel{\text{(\ref{eerr})}}{\approx} (\Tilde{W} - \Phi^T \Phi)_i \Bar{c}(t) 
   \label{waytocompute}
\end{multline}

Then, STDP is applied across each neuron $i$ using $c_i$ as the post- and $\bar{e}$ as the pre-synaptic spike trains for learning $\Phi^T$. STDP is also applied using $f_i$ as the post- and $\bar{e}$ as the pre-synaptic spike trains for learning $W$ \cite{preprint}. Finally, it can be remarked that the SNN-STDP architecture of Fig. \ref{fig1} can naturally be extended to the \textit{convolutional} setting by sweeping the network of Fig. \ref{fig1} across a $H \times W$ image plane with kernel size $[l_x, l_y]$ and stride $d$ \cite{preprint}. A spiking tensor of dimension:
\begin{equation}
    [D_x, D_y, D_c] = [\frac{H-l_x}{d}+1, \frac{W-l_y}{d}+1, M]
    \label{dimension}
\end{equation}
is obtained. In this convolutional case, (\ref{lassodefdivc}) can keep a similar form by considering $\Phi$ as a \textit{circulant} matrix and $\bar{s}$, $\bar{c}$, $\bar{e}$ as flattened tensors along the image plane dimension.
\section{Proposed methods}
\label{methods}
\subsection{Extension to anti-Hebbian learning}
\label{snnstdpbase}
Anti-Hebbian learning (or negative STDP, see Fig. \ref{lifconcept} b) refers to the application of the \textit{opposite} of the STDP rule (\ref{stdppp}) with \textit{negative} $A_+$ and $A_-$ quantities in (\ref{stdppp}), and has been investigated in a few works \cite{asilomar}. Following the derivations in \cite{preprint}, it can be shown that using \textit{negative STDP} for learning the input weights $\Phi^T$ and feedback weights $\Phi$ (see Fig. \ref{fig1}), while using \textit{standard STDP} for the learning of the lateral weights $W$ (\ref{infer}), the following \textit{joint} optimisation objective is solved:
\begin{equation}
 \begin{cases}
 \Bar{c} = \arg \min_{\Bar{c}} ||\Phi \Bar{c} - \Bar{s}||_{2}^{2} + \lambda_1 ||\Bar{c}||_1 %\textbf{ (inference)} 
 \\
 \Phi = \arg \min_{\Phi} -\log (||\Phi \Bar{c} - \Bar{s}||_{2}^{2}) + \frac{\lambda_2}{2} ||\Phi||_F^2 %\textbf{ (learning)}
 \end{cases}
    \label{negativeopt}
\end{equation}
Similar to (\ref{lassodefdivc}), the SNN iterations infer the next SNN output $\bar{c}$ by considering the current $\Phi$ constant and learns the next SNN weight matrix $\Phi$ by considering the current $\bar{c}$ fixed. 

Negative STDP solves (\ref{negativeopt}) since:
\begin{equation}
     \frac{\partial }{\partial \Phi} \{-\log (||\Phi \Bar{c} - \Bar{s}||_{2}^{2}) \} = -\frac{(\Phi \Bar{c} - \Bar{s}) \Bar{c}^T}{||\Phi \Bar{c} - \Bar{s}||_{2}^{2}} 
    \label{antigrad}
\end{equation}
and since it has been shown in \cite{preprint} that \textit{standard} STDP approximates the gradient-based learning:
\begin{equation}
    \Phi \xleftarrow{} \Phi - \eta_2 (\Phi \Bar{c} - \Bar{s}) \Bar{c}^T - \eta_2 \lambda_2 \Phi
\end{equation}
it is now clear that \textit{negative} STDP approximating:
\begin{equation}
    \Phi \xleftarrow{} \Phi - \eta_2 \{-(\Phi \Bar{c} - \Bar{s}) \Bar{c}^T \} -  \eta_2 \lambda_2 \Phi
\end{equation}
is using a gradient similar to (\ref{antigrad}) (up to a dividing scalar oscillating around $\sim 100$ during our experiments) and is learning $\Phi$ in (\ref{negativeopt}). When combined with a \textit{supervision term} (see Section \ref{supervisi}), the use of the \textit{negative} STDP rule can be motivated as a \textit{regularization} controlling the over-fitting of the SNN-STDP by preventing the re-projection $||\Phi \Bar{c} - \Bar{s}||_{2}^{2}$ from becoming too small (see experiments in Section \ref{antistd}).% The effectiveness of this method will experimentally be demonstrated in Section \ref{antistd}.

%In this section, the theoretical tools and the baseline SNN-STDP architecture described in Fig. \ref{fig1} will be used as the basis for the design of our CL system.
%\subsection{Envisioned scenario}
\subsection{Proposed bio-inspired SNN-STDP architecture}
Fig. \ref{figarch2} a) shows the CL SNN-STDP system used throughout this work to infer attention maps. We set up a convolutional network by cascading the SNN-STDP network of Section \ref{snnstdpbase} (in a convolutional setting with \textit{negative} STDP learning) with a convolutional readout in order to learn attention maps. At the output of the convolutional readout, we use \textit{sigmoid} activation functions in order to constrain the dynamic range of the attention map between $0$ and $1$. The rate of the spiking activity at the SNN output is estimated using a rolling time window of length $N_{s} = 20$. Doing so, a tensor $T_{x,y,c}$ containing the mean spike rates for each spatial coordinate $[x,y]$ and each channel $c$ is obtained. Then, $T_{x,y,c}$ is normalized along the channel dimension $c$ to provide invariance to event density \cite{preprint}:
\begin{equation}
    \Tilde{T}_{x,y,c} = \frac{|T_{x,y,c}|}{\sum_c |T_{x,y,c}|}
    \label{norm}
\end{equation}

In between the SNN-STDP layer and the readout, we use \textit{standard scaling} along the channel dimension $c$ in order to reduce the internal covariate shifts \cite{batchnorm}: 
\begin{equation}
    T^*_{x,y,c} = \frac{\Tilde{T}_{x,y,c} - \mu_c}{\sigma_c}
    \label{norm2}
\end{equation}
where the mean $\mu_c$ and standard deviation $\sigma_c$ are estimated on-line. We note the transformations (\ref{norm})(\ref{norm2}) of $\bar{c}$ into $T^*$ as $T^*\{\bar{c}\}$. %following \cite{statestam}. 
Finally, $T^*_{x,y,c}$ is \textit{zero-padded} and fed to the readout layer with a convolutional kernel size $[r_x, r_y, M]$ in order to obtain an attention map $A_{x,y}$ of dimension $[D_x,D_y]$ following (\ref{dimension}). Finally, we interpolate the attention map $A_{x,y}$ back to the input size $[H,W]$.

Without the use of the normalization mechanisms (\ref{norm})(\ref{norm2}), we observed the CL process to fail due to the severe covariate shifts encountered during learning. In addition, we observed that using a convolutional readout layer over a standard fully-connected layer is crucial in order to avoid readout \textit{overfitting} on the local context encountered by the drone during the CL process (CL fails when using a fully-connected layer).
The architecture of Fig. \ref{figarch2} a) also shows that the readout generates a vector of \textit{task-driven} (TD) spike trains $\bar{u}(t)$ that is fed back to the SNN neurons. The TD spike trains are used to \textit{steer} the unsupervised STDP-driven weight dynamics in order to \textit{jointly} optimize the SNN weights $\Phi$ and the readout $\Psi$, forming a task-driven, semi-supervised learning system.

\subsection{Task-driven semi-supervised learning}
\label{supervisi}
During our experiments, we observe that the SNN-STDP system and the readout should not be optimised independently as in \cite{preprint} (i.e., \textit{unsupervised} learning of SNN and \textit{disjoint} learning of a readout). Rather, the SNN and the readout must be learned concurrently. Indeed, if the learning process would be independent, the way the readout should fit the SNN output in the beginning of the STDP learning process will not be the same as in the end, since the SNN weights have \textit{independently} changed since then, leading to significant covariate shifts at the input of the readout%(what the readout has learned at the beginning will not be valid anymore)
. Following this discussion, and the effectiveness of unsupervised DVS learning \cite{preprint}, we opt for a \textit{joint} semi-supervised SNN-readout objective: % where the network infers the next , task-driven dictionary learning approach for joint SNN-STDP and readout learning objective \cite{taskdriven}:
\begin{equation}
\begin{cases}
\Bar{c} = \arg \min_{\Bar{c}} ||\Phi \Bar{c} - \Bar{s}||_{2}^{2} + \lambda_1 ||\Bar{c}||_1 %\textbf{ (SNN inference)}
\\
 \Phi = \arg \min_{\Phi}  \lambda_s \mathcal{L}_{sup}( \bar{\sigma}(\Psi, T^*\{\Bar{c}\}, \bar{b}), \Bar{y}) \\ \hspace{5pt} -\lambda_u \log (||\Phi \Bar{c} - \Bar{s}||_{2}^{2})  + \frac{\lambda_2}{2} ||\Phi||_F^2 %\textbf{ (learns SNN)} 
 \\
\Psi = \arg \min_{\Psi} \lambda_s \mathcal{L}_{sup}( \bar{\sigma}(\Psi, T^*\{\bar{c}\}, \bar{b}), \Bar{y}) %\textbf{ (learns readout)}
\end{cases}
   \label{taskdriveq} 
\end{equation}
where in addition to (\ref{negativeopt}), $\Psi$ is the $K\times M$ weight matrix of the readout, $\bar{b}$ the learned bias vector, $\sigma(.)$ the sigmoid activation, $\bar{y}$ the desired output label vector of dimension $K\times 1$, $\mathcal{L}_{sup}$ the supervised loss function and $\lambda_{u,s}$ the hyper-parameters controlling the strength of the unsupervised and supervised contributions. As usual, the network infers the next $\bar{c}$ keeping $\Phi, \Psi$ constant, then learn $\Phi$ keeping $\Psi, \bar{c}$ constant and so on.
% (\ref{sigmoid})
% \begin{equation}
%   \bar{\sigma}(\Psi, \Bar{c}) = \frac{1}{1 + e^{-(\Psi \bar{c} + \bar{b})}}
%   \label{sigmoid}
% \end{equation}
We choose $\mathcal{L}_{sup}$ as the \textit{focal loss} because of its robustness to imbalanced data compared to the standard cross-entropy \cite{focaloss}:
\begin{equation}
    \mathcal{L}_{sup}(\sigma_i(\Psi_i, T^*\{\Bar{c}\}, \bar{b}),y_i) = -(1-p_i)^\gamma \log(p_i)
    \label{foc1}
\end{equation}
with $\gamma$ setting the robustness to class imbalance and:
\begin{equation}
    p_i= 
    \begin{cases}
    \sigma_i(\Psi_i, T^*\{\Bar{c}\}, \bar{b}) \hspace{3pt} \text{\textbf{if}} \hspace{3pt} y_i =1
    \\
    1-\sigma_i(\Psi_i, T^*\{\Bar{c}\}, \bar{b}) \hspace{3pt} \text{\textbf{otherwise}}
  \end{cases}
  \label{foc2}
\end{equation}

 In order to optimise $\mathcal{L}_{sup}$ as a function of the SNN weights $\Phi$, the link between $\bar{c}$ and $\Phi$ must be made explicit. Since $\bar{c}$ is the result of an iterative process involving the LIF neuron non-linearity, an analytical function $ \bar{c}(\Phi, \bar{s}) $ is hard to find and further relaxations must be done. First, we observe in (\ref{taskdriveq}) that for a small $\lambda_1$, $ \bar{c}(\Phi, \bar{s})$ is well approximated as:
 \begin{equation}
     \bar{c}(\Phi, \bar{s}) \sim (\Phi^T \Phi)^{-1} \Phi^T \bar{s}
     \label{pseu}
 \end{equation}
 
Since $\Phi$ is initialized following an i.i.d. zero-mean normal distribution with standard deviation $\sigma_w=0.01$ \cite{preprint}, the following relaxation holds during the early SNN-STDP learning steps (when most of the learning effect takes place):
 \begin{equation}
      \bar{c}(\Phi, \bar{s})  \sim \frac{1}{N \sigma_w^2} \Phi^T \bar{s}
     \label{relax}
 \end{equation}
 
 Then, the gradient of $\mathcal{L}_{sup}$ as a function of $\Phi$ is:
 \begin{equation}
     \frac{\partial \mathcal{L}_{sup}}{\partial \Phi} = k \bar{s} (\frac{\partial \mathcal{L}_{sup}}{\partial \Bar{c} })^T 
     \label{finalparts}
 \end{equation} 
%  \begin{equation}
%      \frac{\partial \mathcal{L}_{sup}}{\partial \Phi} = k \bar{s} (\frac{\partial \mathcal{L}_{sup}}{\partial \bar{\sigma} })^T \Psi
%      \label{finalparts}
%  \end{equation}
 where $k=\frac{1}{N \sigma^2}$ is a constant (dropped further on) and the gradient %$\frac{\partial \mathcal{L}_{sup}}{\partial \bar{\sigma} }$ 
 $\frac{\partial \mathcal{L}_{sup}}{\partial \Bar{c} }$ directly follows from (\ref{norm})(\ref{norm2})(\ref{foc1})(\ref{foc2}).

\subsection{Task-driven SNN-STDP topology}
Thanks to the relaxation (\ref{relax})(\ref{finalparts}), the original unsupervised SNN-STDP architecture of Fig. \ref{fig1} can now be modified in order to incorporate the effect of the newly added task-driven term in (\ref{taskdriveq}). We observe in (\ref{finalparts}) that:
\begin{equation}
    \bar{v} = (\frac{\partial \mathcal{L}_{sup}}{\partial \Bar{c} })^T %(\frac{\partial \mathcal{L}_{sup}}{\partial \bar{\sigma} })^T \Psi
    \label{taskdrivcont}
\end{equation}
is a $1\times M$ vector computed at the \textit{readout side} (e.g., in an MCU used to post-process the SNN activity) while the \textit{spike train} vector $\bar{s}(t)$ in (\ref{finalparts}) is already available in both the \textit{coding} and \textit{error} layers (see Fig. \ref{fig1} a). Therefore, the TD contribution can be injected in the SNN-STDP system by first converting $\bar{v}$ to a push-pull pair of spike trains as follows:
\begin{equation}
    \bar{u}(t) = LIF_{pp}\{\bar{v}\} %- LIF \{ -\bar{v} \}
    \label{epsig}
\end{equation}
%where $LIF\{.\}$ denotes the application of the LIF non-linearity (\ref{liff}). 
Then, each entry $u_i$ of $\bar{u}(t)$ can be fed to the $i^{th}$ neuron in the \textit{coding} layer via one additional synapse (see Fig. \ref{figarch2} b), and the spiking vector $\bar{u}(t)$ can be distributed to each \textit{error} neuron via an additional set of synapses (see Fig. \ref{figarch2} c). 

Finally, in addition to the unsupervised STDP mechanisms already present in the network of Fig. \ref{fig1} to solve (\ref{lassodefdivc}), it follows from (\ref{approx}) that an additional STDP contribution must be applied to the neurons in the \textit{error} layer as follows:
\begin{equation}
    (\Phi_j)_i \xleftarrow{} (\Phi_j)_i - \lambda_s \Delta w_{\text{STDP}} \{s_j(t), u_i(t)\}
    \label{firstlololffff}
\end{equation}
with $s_j(t)$ the post- and $u_i(t)$ the pre-synaptic spike trains. An additional STDP contribution must also be applied to the neurons in the \textit{coding} layer as:
\begin{equation}
    (\Phi_i^t)_j \xleftarrow{} (\Phi_i^t)_j - \lambda_s \Delta w_{\text{STDP}} \{u_i(t), s_j(t)\}
    \label{stdprew1}
\end{equation}
with $u_i(t)$ the post- and $s_j(t)$ the pre-synaptic spike trains. Regarding the lateral weights $W$, it can be shown \cite{preprint} that the STDP mechanism adopted in the network of Fig. \ref{fig1} automatically enforces consistency between the task-driven learning of $\Phi$ and the convergence of $W$ (since applying STDP between $f_i(t)$ in (\ref{waytocompute}) and $\bar{e}(t)$ in (\ref{eerr}) forces $W$ to converge to $\eta_1 \Phi^T\Phi -I_M$ in (\ref{infer}) \cite{preprint}).   

\subsection{Continual learning strategy}
\label{wherethresh}
For on-line labelling, we use a pre-trained YOLOv3 network \cite{yolov3} to detect the presence or absence of a human subject from the RGB frames that are jointly acquired with the DVS data, and to infer the bounding box coordinates $[b_{x,1}, b_{y,1}, b_{x,2}, b_{y, 2}]$ (normalized to the size of the output map) indicating the location of the human subject in the image. When a human subject is present, the corresponding \textit{label map} $\bar{y}$ (used for the supervised contribution) is assigned with \textit{ones} inside the bounding box region and \textit{zeros} outside.
% \begin{equation}
% \begin{cases}
%     \bar{y}[\frac{b_{x,1} + b_{x,1}}{2}, \frac{b_{y,1} + b_{y,1}}{2}] = 1
%     \\
%     \bar{y}[x, y] = 0 \hspace{10pt} \forall x \neq \frac{b_{x,1} + b_{x,1}}{2}, \hspace{5pt} \forall y \neq \frac{b_{y,1} + b_{y,1}}{2}
% \end{cases}
% \end{equation}
Since there are significantly more zero-valued entries in $\bar{y}$, we use \textit{median frequency balancing} \cite{medianfreq} to help mitigate the learning imbalance in space. Therefore, we weight the readout loss as follows:
\begin{equation}
    \mathcal{L}_{sup}(\sigma_i,y_i) \xleftarrow{} w_i \mathcal{L}_{sup}(\sigma_i,y_i) \hspace{5pt} \textbf{with} \hspace{5pt} w_i = \frac{f_m}{N_{class,i}}
\end{equation}
where $N_{class,i}$ is the number of entries in $\bar{y}$ of the same class as the $i^{th}$ label entry (the class can be either $0$ or $1$ in our detection case) and $f_m$ is the mean of $N_{class = 0}$ and $N_{class=1}$.

When no human subjects are detected, the label map $\bar{y}$ will only contain null values for long iteration periods, leading to the \textit{over-fitting} of the system on this local context. We alleviate this problem by keeping $\Psi$ \textit{fixed} when no human subjects are present, and only enabling learning for $\Phi$. In addition, we \textit{prune} the task-driven contribution to the learning of $\Phi$ (\ref{taskdrivcont}) for values of $\mathcal{L}_{sup}$ \textit{smaller} than a threshold $\theta_{th}=0.2$ (empirically found to work well). Doing so, we avoid the over-fitting of the system by sporadically learning hard examples only.

Regarding the choice of the on-line optimizer for the convolutional readout, we choose the \textit{Adam} optimizer \cite{adam} because of its adaptive gradient scaling capability, which, in an on-line learning setting, leads to faster convergence compared to SGD. In addition, the use of momentum in Adam implicitly incorporates gradients from the data in hind-side, re-balancing the learning process over time \cite{adam}. 
\subsection{Post-processing}
\label{postprocess}
A set of discrete detections is obtained from the attention map as follows. First, a threshold $d_{th}$ is applied to the map and a point cloud is formed. Then, a set of detected clusters $\{C_k\}$ is obtained by applying DBSCAN clustering \cite{dbscan} with \textit{min points} $=2$ and $\epsilon = 5$ (empirically tuned). 
\section{Experimental results}
\label{expres}
In this section, we assess the bio-inspired CL architecture described in Section \ref{methods} on the \textit{KUL-UAVSAFE} dataset \cite{ali} for people detection on drones. The \textit{KUL-UAVSAFE} dataset features a collection of joint DVS and RGB acquisitions in an \textit{indoor}, industrial-like environment, where a human subject is walking randomly. During our experiments, we choose the three longest acquisitions ($\sim 2$ min.) from the \textit{f-wall} collection in \cite{ali}, where the drone and a human subject are moving in a space surrounded by walls, benches and shelves (different human subjects are featured to add variability). 

We assess the performance of our system via 3-fold cross-validation as follows. First, one of the acquisitions is used as a \textit{learning sequence} and is shown to our CL system \textit{only once}, in its natural order (the learning sequence \textit{is not shuffled}). Then, learning is stopped and \textit{a different acquisition} is used to measure the \textit{precision-recall} curves after post-processing (Section \ref{postprocess}), by sweeping $d_{th}$ from $0$ to $1$ and measuring the number of false alarms, true positives and false negatives in $\{C_k\}$. This process is repeated three times using different learning and testing sets and the final precision-recall curves are obtained as the average over the three learn-test folds. 
\begin{figure}[htbp]
\centering
    \includegraphics[scale = 0.57]{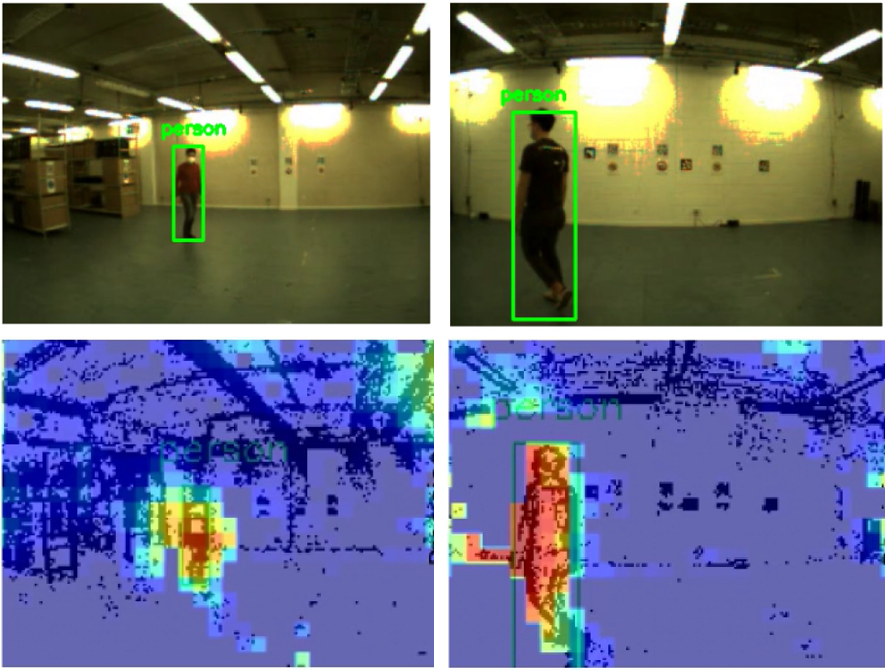}
    \caption{\textit{\textbf{Attention maps} inferred from the DVS data by our bio-inspired CL SNN-STDP system. An off-the-shelve YOLOv3 is used for online labelling. }}
    \label{florilege}
\end{figure}

Fig. \ref{florilege} shows examples of attention maps produced by our system. Table \ref{paramsdvstrain1} reports the SNN-STDP learning parameters (found empirically, starting from the parameter set used in \cite{preprint}). Table \ref{paramsdvstrain2} reports the parameters used in our system architecture (tuned empirically). For the readout, we use the default values of the Adam optimizer \cite{adam} with a learning rate $\eta_r=5\times 10^{-6}$. Next, we will study the effect of the most important parameters impacting the CL procedure: the focal loss parameter $\gamma$ (\ref{foc1}) and the semi-supervision hyper-parameters $\lambda_{u,s}$ (\ref{taskdriveq}).% and the loss threshold parameter $\theta_{th}$ (see Section \ref{wherethresh}).
%dbscan
%params
\begin{table}[htbp]
\centering
\begin{tabular}{|c|c|c|c|c|c|c|c|c|}
\hline
$\eta_1|\eta_2$ & $\mu_1|\tau_m$ & $\lambda_2$ & $A_+$ & $A_-$ & $\tau_+$ & $\tau_-$ & $\tau_s$\\
\hline
$0.07|0.05$ & $0.15|0.07$ & $0.002$ & 1 & 0.8 & 0.02 & 0.008 & 0.01\\
\hline
\end{tabular}
\caption{\textit{\textbf{SNN-STDP learning parameters.} The simulation time step is $0.005$s. All $\tau$ values are in seconds. $\lambda_1$ in (\ref{taskdriveq}) is implicitly set by the neuron threshold $\mu$ \cite{preprint}. $\theta_{th} = 0.2$. The negative STDP parameters are $A_{+,-}^n = A_{+,-}/100$. }}
\label{paramsdvstrain1}
\end{table}
\begin{table}[htbp]
\centering
\begin{tabular}{|c|c|c|c|c|c|c|c|}
\hline
$H$ & $W$ & $l_{x,y}$ & $d$ & $M$ & $r_{x,y}$ & $D_x$ & $D_y$\\
\hline
130 & 173 & 30 & 5 & 64 & 12 & 21 & 29 \\
\hline
\end{tabular}
\caption{\textit{\textbf{Architectural parameters.} $H \times W$: input size. $l_x \times l_y$: SNN kernel size, $d$: SNN stride, $M$: number of SNN kernels (neurons in the coding layer), $r_x \times r_y$: kernel size of the readout. $D_x \times D_y$: shape of the output attention map. The readout has a stride of 1 and zero-padding.}}
\label{paramsdvstrain2}
\end{table}
\subsection{Impact of the focal loss parameter $\gamma$}
\begin{figure}[htbp]
\centering
    \includegraphics[scale = 0.57]{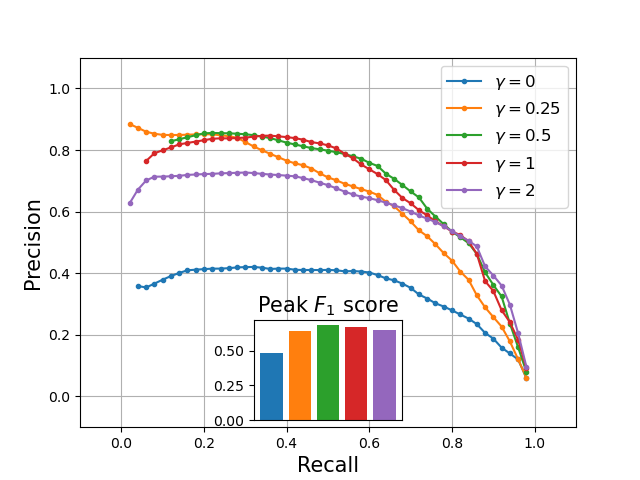}
    \caption{\textit{\textbf{Impact of the focal loss parameter $\gamma$}. The peak $F_1$ score is maximized near $\gamma=0.5$.}}
    \label{choosegamma}
\end{figure}

We explore the impact of $\gamma$ in (\ref{foc1}) by keeping all other parameters constant with $\lambda_u=0$, $\lambda_s=1$. Fig. \ref{choosegamma} shows the measured precision-recall curves. Fig. \ref{choosegamma} also reports the peak $F_1$ score as $\max_i \frac{2 P_i R_i}{P_i + R_i}$ along each precision-recall curve $(P_i,R_i)$ \cite{ali}. Fig. \ref{choosegamma} shows that the peak $F_1$ score is maximised near $\gamma = 0.5$. Therefore, $\gamma = 0.5$ in our experiments below. 
%\subsection{Impact of the coding neuron number $M$}
%Therefore, we choose $M = $ \textcolor{blue}{put} in the rest of our experiments.
\subsection{Impact of the unsupervised and supervised strengths $\lambda_u, \lambda_s$}
\label{antistd}
\begin{figure}[htbp]
\centering
    \includegraphics[scale = 0.57]{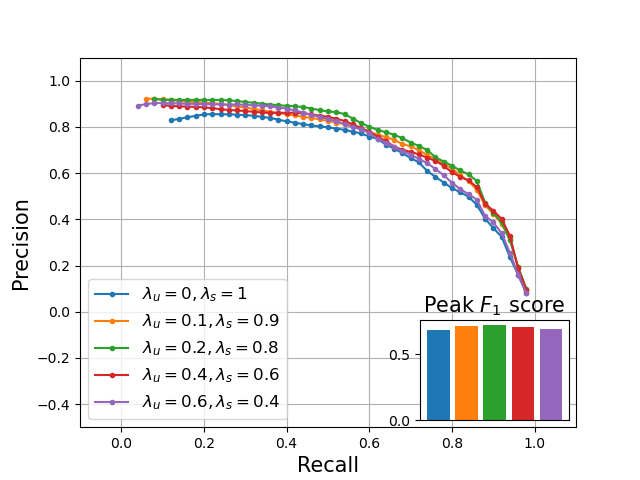}
    \caption{\textit{\textbf{Varying the semi-supervised hyper-parameters $\lambda_{u,s}$}. The peak $F_1$ score is maximized near $\lambda_u=0.2$ and $\lambda_s=0.8$.}}
    \label{chooselamu}
\end{figure}

Fig. \ref{chooselamu} shows the precision-recall curves obtained by varying the \textit{unsupervised} (negative STDP) and \textit{supervised} contributions $\lambda_{u,s}$ in (\ref{taskdriveq}). Fig. \ref{chooselamu} shows that the peak $F_1$ score is maximised near $\lambda_u=0.2$, $\lambda_s=0.8$.
During our experiments, we also tried the use of the standard \textit{positive} STDP rule for the unsupervised contribution, but always reached lower performances by varying $\lambda_u$, due to an increase in false alarms. In contrast, the use of the \textit{negative} STDP rule for the \textit{unsupervised} contribution provides an \textit{opposite weight-steering force} to the \textit{positive} STDP rule used for the \textit{supervised} contribution, preventing the over-fitting of the system to structures in the environment. This leads to less false alarms at \textit{higher recalls} and therefore, a higher detection performance (a non-trivial gain of +$5\%$ on the peak $F_1$ score vs. $\lambda_u=0$, $\lambda_s=1$). 
\subsection{Continual SNN-STDP vs. standard CNNs}
\begin{figure}[htbp]
\centering
    \includegraphics[scale = 0.57]{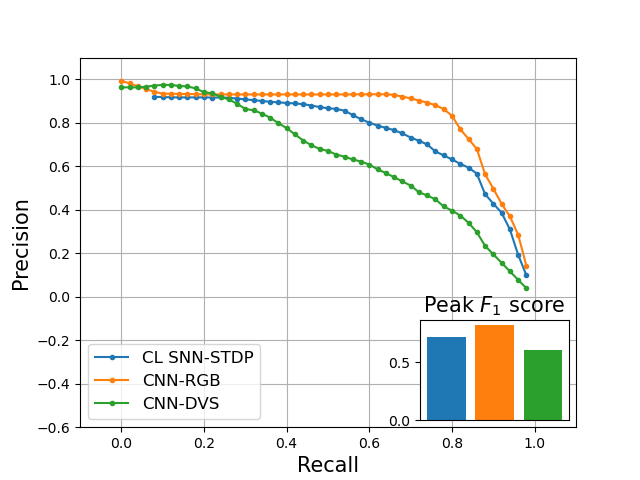}
    \caption{\textit{\textbf{Detection precision vs. recall} for the presented CL SNN-STDP and the conventional CNNs with offline training.}}
    \label{comparefin}
\end{figure}

Since CNNs constitute the \textit{standard approach} used in the vision pipelines of drones \cite{computematter}, we compare in Fig. \ref{comparefin} the performance of our CL architecture to two offline-trained CNNs: a SqueezeNet-based CNN trained with RGB data and a second SqueezeNet trained with DVS \textit{frames} following \cite{ali} (DVS data averaged over time into frames at 30 FPS). 

%three offline-trained systems: the SNN system of this work \textit{trained on a shuffled dataset} (everything else kept identical) and a SqueezeNet-based CNN trained with either RGB or DVS \textit{frames} following \cite{ali} (DVS data averaged over time into frames at 30 FPS).
Using a SqueezeNet CNN architecture is an appropriate choice for comparison since SqueezeNet aims to be MCU-friendly, using $\sim500$KB of weight memory (comparable to our network that uses $514$KB). In order to \textit{fairly} compare the bounding box output of the CNN to the attention map of our SNN-STDP system, we consider the image region delimited by the bounding box as the active region of an attention map (region inside the bounding box is filled with $1$ and outside is filled with $0$). We can then use the same precision-recall measurement pipeline as used for the SNN. %Note that this is in the advantage of the CNN since using a typical \textit{intersection over unions} (IoU) threshold \cite{ali} is more restrictive and leads to lower precision-recall curves.   
Fig. \ref{comparefin} shows that the CNN-RGB setup is the best performer (due to less false alarms compared to our SNN), but this comes at the cost of \textit{on-line adaptability}. Indeed, the CNN has been trained offline for a specific environment and needs re-training each time new situations are encountered. This is in contrast to our CL SNN-STDP system which can learn and adapt on the fly, as demonstrated in this work. Interestingly, the CNN-DVS setup is the lowest performer and is significantly outperformed by the CL SNN-STDP system (+19\% on the peak $F_1$ score). This is mainly due to the fact that the SNN introduces \textit{time recurrence} through its lateral weights, enabling the network to learn \textit{temporal features} that are neglected by the feed-forward CNN-DVS (see \cite{hots} for a discussion on temporal features in DVS data). %It must be noted that but still, the extension of those methods to on-line continual learning is unclear. %Finally, the offline-trained SNN-STDP setup (shuffled dataset) \textit{only marginally} outperforms the CL system, clearly showing the effectiveness of the methods proposed in Section \ref{methods}.%for learning \textit{on the fly}. 
%frame dvs vs event by event, but cnn cannot learn and adapt online, contrary to what we have demonstrated in this paper inne recurrence of the model vs cnn
This clearly shows the effectiveness of the methods proposed in Section \ref{methods} for the continual learning of DVS data.
\section{Conclusion}
\label{conc}
This paper has presented, to the best of our knowledge, the first continual learning system for drones that learns to detect people on the fly using a bio-inspired SNN architecture with STDP learning. After the introduction of our novel methods, numerous experiments have been described to characterize the performance of our system and compare it against conventional CNNs. We have shown that our event-based system reaches a higher peak $F_1$ score (+19\%) compared to a same-size CNN processing DVS frames, while enabling on-line adaptation and learning on the fly. As future work, we plan to increase the precision of the system at higher recalls by studying how to enhance the SNN expressivity with more layers and reducing false alarms. We hope that this work will inspire future research in brain-inspired vision for robotics. 

\section*{Acknowledgment}
We thank Lars Keuninckx and Tim Verbelen for their precious help.
The research leading to these results has received funding from the Flemish Government (AI Research Program) and the European Union's ECSEL Joint Undertaking under grant agreement n° 826655 - project TEMPO.

\end{document}